\pdfoutput=1

\documentclass[11pt]{article}

\usepackage[]{acl}

\usepackage{times}
\usepackage{latexsym}

\usepackage[T1]{fontenc}

\usepackage[utf8]{inputenc}

\usepackage{microtype}

\usepackage{inconsolata}

\usepackage{booktabs}
\usepackage{comment}

\usepackage{graphicx}
\usepackage{caption}
\usepackage[normalem]{ulem}
\usepackage{multirow}
\useunder{\uline}{\ul}{}

\usepackage{enumitem}
\usepackage{tabularx}
\usepackage[export]{adjustbox}
\usepackage{hyperref}
\usepackage{amsmath}
\usepackage{graphicx}
\usepackage{bbm}
\usepackage{tablefootnote}

\newcommand{\lin}[1]{{\textcolor{orange}{#1}}}

\usepackage{hyperref} 

\title{PreFLMR: Scaling Up Fine-Grained Late-Interaction Multi-modal Retrievers}

\author{\thanks{Weizhe Lin, Jingbiao Mei, and Jinghong Chen equally contributed to this work.}Weizhe Lin, Jingbiao Mei, Jinghong Chen, Bill Byrne \\
  Department of Engineering \\
  University of Cambridge \\
  Cambridge, United Kingdom CB2 1PZ \\
  \texttt{\{wl356, jm2245, jc2124, wjb31\}@cam.ac.uk}}

\begin{document}
\maketitle
\begin{abstract}
Large Multimodal Models (LMMs) excel in natural language and visual understanding but are challenged by exacting tasks such as Knowledge-based Visual Question Answering (KB-VQA) which involve the retrieval of relevant information from document collections to use in shaping answers to questions.    
We present an extensive training and evaluation framework, M2KR, for KB-VQA.    
M2KR contains a collection of vision and language tasks which we have incorporated into a single suite of benchmark tasks for training and evaluating general-purpose multi-modal retrievers. 
We use M2KR to develop PreFLMR, a pre-trained version of the recently developed  Fine-grained Late-interaction Multi-modal Retriever (FLMR) approach to KB-VQA, and we report new state-of-the-art results across a range of tasks. 
We also present investigations into the scaling behaviors of PreFLMR intended to be useful in future developments in general-purpose multi-modal retrievers.
The code, demo, dataset, and pre-trained checkpoints are available at \href{https://preflmr.github.io/}{https://preflmr.github.io/}.
\end{abstract}

\section{Introduction}
\label{sec:preflmr:introduction}

Knowledge-based Visual Question Answering (KB-VQA) systems generate answers to queries consisting of questions about given images. Correctly answering these questions requires accessing relevant world knowledge as well as vision and language understanding.  Despite their demonstrated abilities in vision and language,  recent Large Multimodal Models (LMMs)~\cite{chen2023pali, driess2023palme, liu2023improved, zhu2023minigpt, openai2023gpt4} have performed poorly in recent challenging KB-VQA tasks~\cite{chen-etal-2023-pre-trained, Mensink_2023_ICCV}.  One promising approach to improve their KB-VQA performance is Retrieval-Augmented Generation (RAG), in which answer generation by LMMs is grounded in relevant documents retrieved from a knowledge base.

The best-performing document retrieval approach for KB-VQA to date is Fine-grained Late-interaction Multi-modal Retrieval (FLMR)~\cite{lin2023finegrained}. FLMR uses multi-dimensional embedding matrices to represent documents and queries and then efficiently computes their relevance scores via late-interaction~\cite{khattab-zaharia-2020-colbert}, thus capturing fine-grained relevance at the token level rather than at the passage level, as in Dense Passage Retrieval (DPR)~\cite{karpukhin-etal-2020-dense}. 
As a late-interaction retriever, FLMR substantially outperforms DPR on a range of KB-VQA tasks, with only minor speed penalties. 
In all of these methods, model and data size are important considerations. 
There has been much work in scaling up Large Language Models (LLMs)~\cite{kaplan2020scaling, NEURIPS2022_8c22e5e9, chen2023pali} and text-based retrieval~\cite{ni-etal-2022-large}, 
but the scaling properties of these vision and language retrieval systems have not been studied. 
We therefore investigate the following three aspects of FLMR in KB-VQA. 

(1) \emph{Vision \& Text Encoding}:   We investigate how KB-VQA performance is affected by scaling the size and complexity of vision and text encoders.

(2) \emph{Pre-training}:  
As originally formulated, FLMR employs simple, lightly trained Multi-Layer Perceptrons (MLP).  We investigate whether gains can be had through more extensive model pre-training. 



(3) \emph{Task Diversity}:  We gather nine open-source vision-language datasets into a suite of benchmark tasks, M2KR, for assessing Multi-task Multi-modal Knowledge Retrieval.   M2KR encompasses Image-to-Text, Question-to-Text, and Image\&Question-to-Text retrieval tasks, and also includes prompting instructions that can be provided to an LLM for each of the component tasks.  General purpose multi-modal retrieval models can be created by training on the entirety of the M2KR training data and these models can then be evaluated on any or all of the included tasks.  Models can further be fine-tuned for specific M2KR tasks using the task-specific tuning data included in the collection.  

We show that M2KR can be used in training an FLMR-based RAG LLM for multi-task multi-modal retrieval.   We refer to this model as PreFLRM (for Pre-trained FLMR).  PreFLMR can be used directly in its pre-trained form for multi-task multi-modal retrieval.  PreFLMR can also be fine-tuned for specific task-specific performance.  In both uses we find that PreFLMR gives us substantial gains across the M2KR tasks.  

Contributions of this paper are:
\begin{itemize}[leftmargin=5.5mm]
\setlength\itemsep{-0.3em}
\item The M2KR task suite encompassing nine datasets and three types of retrieval tasks for training and evaluating general-purpose vision-language retrievers. We create M2KR by re-purposing various vision and language data sets that might not be originally created for knowledge-based visual question answering, thus ensuring a rich and diverse collection.
\item PreFLMR, a strong multi-modal retriever pre-trained on a vision-language corpus of over ten million  items.  We show that PreFLMR performs well across a range of knowledge retrieval tasks when given the appropriate instructions. We will release PreFLMR upon publication. 
\item A study of the scaling behaviour of FLMR in terms of its model parameters and training data. To our knowledge, this is the first systematic study of scaling in late-interaction based vision-language retrievers and should provide empirical guidance for future work. 
\end{itemize}

\section{Related Work}

\noindent\textbf{Document Retrieval.}
DPR has become a cornerstone in knowledge-intensive tasks~\cite{chen-etal-2017-reading, izacard-grave-2021-leveraging, guu2020realm, lee-etal-2019-latent, lewis2020retrieval} as well as in KB-VQA tasks due to its fast and precise retrieval capabilities~\citep{karpukhin-etal-2020-dense, gui2021kat, luo-etal-2021-weakly, lin-byrne-2022-retrieval, wu-mooney-2022-entity}. Recent developments in retrieval methods, particularly Late Interaction models~\citep{khattab-zaharia-2020-colbert, santhanam-etal-2022-colbertv2}, have shown notable performance gains over DPR, albeit with some efficiency trade-offs~\cite{lin-etal-2023-li, lin2023finegrained}. In multi-modal retrieval, FILIP~\citep{yao2022filip} used pre-trained late interaction models for single-modal image-text retrieval, while FLMR~\cite{lin2023finegrained} extended the approach to multi-modal retrieval for KB-VQA with finer-grained visual and text features.  This paper further extends FLMR and explores its scaling properties in multi-modal retrieval.
Similar to our M2KR benchmark, A concurrent work~\cite{wei2023uniir} introduces M-Beir, which combines several retrieval tasks and can be used to train and evaluate universal multi-modal retrievers.

Another line of relevant research is KB-VQA retrieval involving Named Entities, where retrieved documents must identify the person in the image. 
For example, on ViQuAE~\cite{lerner2022viquae}, \citet{lerner2023multimodal} trains the retriever with a multi-modal inverse cloze task, while \citet{lerner2024cross} shows that combining mono- and cross-modal retrieval improves performance. Both use a weighted sum of BERT~\cite{devlin-etal-2019-bert} and CLIP~\cite{Radford_2021_CLIP} embeddings, while our work trains a single multi-modal late-interaction retriever.

\noindent\textbf{Knowledge-based VQA Systems.} 
Recent multi-modal systems have significantly improved in complex tasks like OKVQA~\cite{schwenk2022okvqa} that require external knowledge sources~\citep{narasimhan2018out, garderes2020conceptbert, li2020boosting, wu2022multi, marino2021krisp, chen2023lako, gao2022transform, gui2021kat, Hu2023REVEAL, rao2023retrieval}. Systems like KAT~\citep{gui2021kat} and REVIVE~\citep{lin2022revive} used LLMs (e.g. GPT-3) for generating candidate answers. Challenges remain in answering more knowledge-intensive questions~\cite{chen-etal-2023-pre-trained, Mensink_2023_ICCV}, underscoring the need for robust document retrieval.  \citet{Mensink_2023_ICCV} showed that even state-of-the-art LLMs perform poorly on difficult KB-VQA questions, with an accuracy of under 20\% when retrieval is not incorporated. RA-VQAv2~\citep{lin2023finegrained} and prior work~\citep{lin-byrne-2022-retrieval, luo-etal-2021-weakly, qu2021passage, gao2022transform, Hu2023REVEAL, Mensink_2023_ICCV} demonstrated strong performance in KB-VQA by using external knowledge databases.

\noindent\textbf{Scaling Retrieval Systems.}
Previous work has explored scaling laws in language/vision systems~\cite{kaplan2020scaling, NEURIPS2022_8c22e5e9}, revealing correlations between model performance, computation, number of parameters, and dataset sizes. 
In retrieval, \citet{ni-etal-2022-large} and \citet{Hu2023REVEAL} both observe improvements in DPR-like models with one-dimensional embeddings by increasing the size of language/vision encoders.
This paper reports similar scaling investigations in multi-modal late-interaction retrieval.


\section{The M2KR Benchmark Suite}
Current multi-modal retrievers are typically {trained and} evaluated on a single dataset only. To properly study general-purpose multi-modal retrievers, we introduce the Multi-task Multi-modal Knowledge Retrieval (M2KR) benchmark suite. 
We convert nine diverse datasets, originally designed for vision and language tasks such as image recognition, image captioning, and conversational interactions, into a uniform retrieval format.  
Details of the pre-processing steps, data partition, and prompting instructions are provided in Appendix~\ref{sec:preflmr:appendix:dataset_detail}, but we note here that re-purposing these datasets into a single consistent collection for knowledge-based visual question answering represents a non-trivial effort.  M2KR will be released with our models. 

\subsection{Tasks and Datasets}

Table~\ref{tab:dataset_stats} shows the composition of M2KR. We pre-process the datasets into a uniform format and write several task-specific prompting instructions for each dataset.  The M2KR benchmark contains three types of tasks:

\paragraph{Image to Text (I2T) retrieval.} 
These tasks evaluate the ability of a retriever to find relevant documents associated with an input image.   
Component tasks are WIT~\citep{srinivasan2021wit}, IGLUE-en \cite{Bugliarello2022iglue}, KVQA \cite{Shah2019KVQA}, and CC3M~\citep{Sharma2018CC3M}.  
CC3M is included in the M2KR training set to improve scene understanding but not in the validation/test set as the task concerns caption generation, not retrieval.
The IGLUE test set, which is a subset of WIT and has an established benchmark, is included to enable comparison with the literature.
The KVQA task, initially designed as a KB-VQA task, has been re-purposed into an I2T task for our modelling purposes (Appendix~\ref{sec:preflmr:dataset_detail:kvqa}).

\paragraph{Question to Text (Q2T) retrieval.} 
This task is based on MSMARCO \cite{Bajaj_2018MSMARCO} and is included to assess whether multi-modal retrievers retain their ability in text-only retrieval after any retraining for images. 

\paragraph{Image \& Question to Text (IQ2T) retrieval.}
This is the most challenging task which requires joint understanding of questions and images for accurate retrieval.  It consists of these subtasks:  
OVEN~\cite{Hu2023Oven}, LLaVA~\citep{Liu2023llava}, OKVQA \cite{schwenk2022okvqa}, Infoseek \cite{Chen2023infoseek} and E-VQA \cite{Mensink2023EVQA}. 
We note in particular that we convert LLaVA, a multi-modal conversation dataset, into a multi-modal retrieval task (Appendix~\ref{sec:preflmr:dataset_detail:llava}).

The training/validation/test examples are downsampled from the respective sets of the original datasets.
We take test examples from the original validation sets for LLaVA and Infoseek since LLaVA has no test sets and the test set annotation of Infoseek has not been released.
We limit the maximum test samples to 5,120 for each dataset to allow faster performance tests on all 9 datasets.
Data preprocessing and partitioning details are in Appendix~\ref{sec:preflmr:appendix:dataset_detail}.
We further verified that there are no identical images between the training and test sets by checking the MD5 of the images, thereby preventing data contamination during training.
We use the validation splits to select hyperparameters for the models which can be found in detail in Appendix~\ref{sec:preflmr:appendix:hyperparameters}



\begin{table}[]
    \centering
    \resizebox{\columnwidth}{!}{%
    \small
    \begin{tabular}{lrrrrr}
    \toprule
     & \multicolumn{3}{c}{\#Examples} & \multicolumn{2}{c}{\#Passages} \\
    Datasets                             & Train & Val & Test & Train & Val/Test \\
    \midrule
    \multicolumn{4}{l}{\textit{I2T Retrieval}} \\
    \midrule
    WIT         & 2.8M              & 20,102 & 5,120                & 4.1M    &     40K            \\
    \textit{ IGLUE}  & - & -& 685  & -  & 1K \\
    KVQA                   & 65K             & 13,365 &  5,120               & 16.3K            & 4,648       \\
    CC3M                  & 595K            & - &  -               & 595K & -                 \\
    \midrule
    \multicolumn{4}{l}{\textit{Q2T Retrieval}} \\
    \midrule
        MSMARCO                &     400K    & 6,980 &       5,120       &    8.8M    & 200K             \\
    \midrule
    \multicolumn{4}{l}{\textit{IQ2T Retrieval}} \\
    \midrule
    
    OVEN                   & 339K   &   20,000       &  5,120               & 10K  & 3,192                \\
    LLaVA                               & 351K             & - & 5,120                & 351K  & 6,006                \\
    OKVQA                              & 9K              & 5,046 &  5,046               & 110K & 110K                  \\
    Infoseek                            & 100K            & - &  4,708           & 100K  & 100K     \\
    E-VQA                  & 212K            &  9,852  & 3,750             & 50K             & 50K      \\

    \bottomrule
    \end{tabular}
    }
    \caption{Datasets in M2KR Benchmark Suite. 
    }
    \label{tab:dataset_stats}
    \vspace{-0.2cm}
\end{table}

\subsection{Evaluation}
\label{sec:M2KR-evaluation}
We use \textit{Recall@K (R@K)}, which measures whether at least one of the target documents is in the top-$K$ retrieved entries, to evaluate retrieval performance.
Additionally, for the datasets Infoseek, E-VQA, and OKVQA, we mainly employ \textit{Pseudo Recall/PRecall@K (PR@K)} for evaluation. This metric measures whether at least one of the top K documents includes the target answer.\footnote{In practice, PR@K more accurately reflects actual retrieval performance and exhibits a stronger correlation with the ultimate VQA performance. This is because document annotations are frequently incomplete, and alternative documents within the corpus can often provide answers to the questions.}

We use R@10 for WIT and MSMARCO, and R@1 for LLaVA and IGLUE.
Other datasets are evaluated with R@5 or PR@5.
As in Table~\ref{tab:preflmr:main_results}, we also report the average rank (\textit{A.R.}) of each model over all datasets to indicate multi-task retrieval performance relative to other models in comparison; lower is better.



\subsection{Baselines}


For each dataset, we show the best published results in recent literature as baselines,  if available (Table~\ref{tab:preflmr:main_results}). 
For datasets without previous results such as LLaVA and OVEN, we use our replication of CLIP~\cite{Radford_2021_CLIP} and FLMR as zero-shot  baselines following~\citet{lin2022revive}. 


\section{PreFLMR Architecture and Training}
\begin{figure*}
    \centering
    \includegraphics[width=\textwidth]{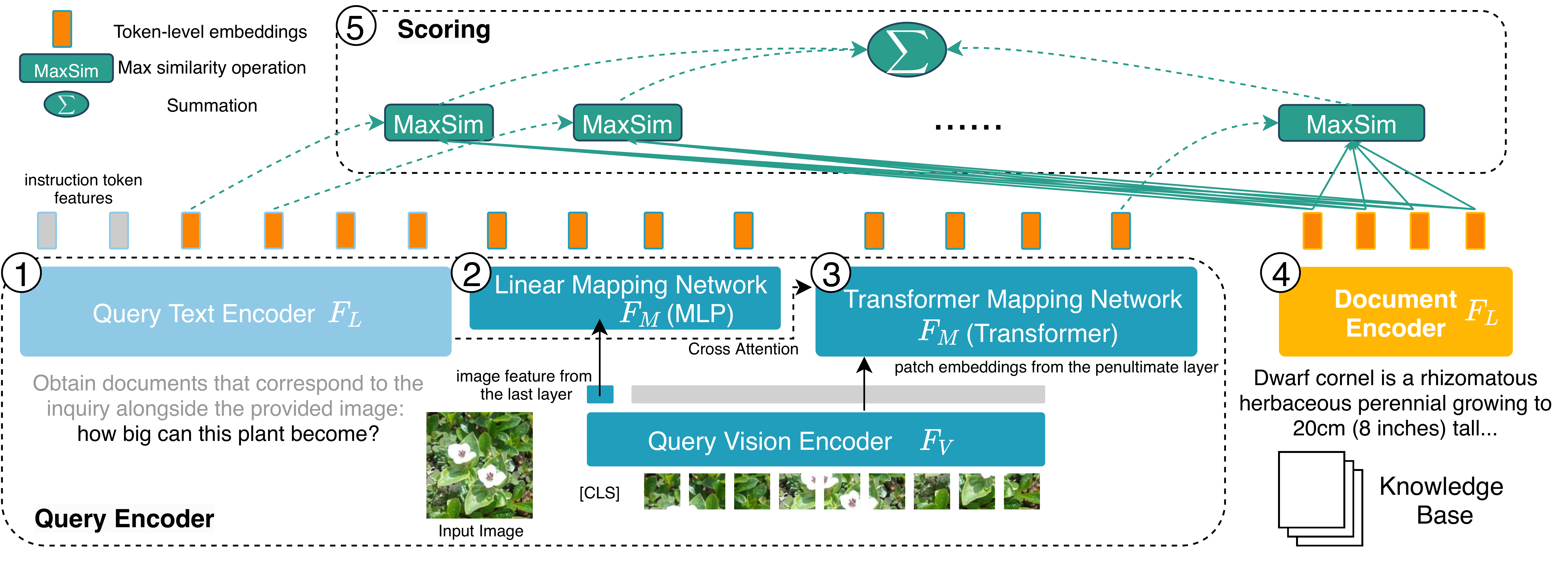}
    \caption{PreFLMR Model Architecture. (1) the text query consists of an instruction and a question, which is encoded by a text encoder; (2) at the output of the vision encoder, a mapping network consisting of Multi-Layer Perceptrons (MLP) converts the `[CLS]' token representations into the same embedding space as the text encoder; (3) the transformer blocks take in the patch image embeddings from the penultimate layer of the vision encoder and attend to the text features by cross-attention; (4) a text encoder encodes documents in the knowledge base; (5) the scores between queries and documents are computed based on late-interaction, allowing each query token to interact with all document token embeddings.}
    \label{fig:preflmr:overview}
\end{figure*}

Our architecture generally follows that of FLMR~\cite{lin2023finegrained} as shown in Fig.~\ref{fig:preflmr:overview}.
PreFLMR uses token embedding matrices $\mathbf{Q}$ and $\mathbf{D}$ to represent query and document, respectively. Given a query $\bar{\textbf{q}}$ consisting of texts $q$ and an image $I$, PreFLMR uses a language model $\mathcal{F}_{L}$ to obtain embeddings of all tokens in $q$, a vision model $\mathcal{F}_{V}$ to obtain embeddings of $I$, and a mapping structure $\mathcal{F}_{M}$ to project image embeddings into the text embedding space. All token-level embeddings are concatenated to form the query representation $\mathbf{Q}$. The document matrix $\mathbf{D}$ is obtained similarly with the language model $\mathcal{F}_{L}$ but without visual features. 

The relevance score $r(\bar{\mathbf{q}}, d)$ is computed via \textit{late-interaction}~\cite{khattab-zaharia-2020-colbert} between $\mathbf{Q}$ and $\mathbf{D}$, aggregating the maximum dot products over all query tokens with respect to all document tokens (Eq.~\ref{eqn:flmr:late_interaction}). $l_Q$ and $l_D$ denote the total number of tokens in query $\bar{\mathbf{q}}$ and document $d$, respectively.

\begin{equation}
r(\bar{\mathbf{q}}, d) = \sum_{i=1}^{l_Q} \max_{j=1}^{l_D} \mathbf{Q}_{i} \mathbf{D}_{j}^\top
\label{eqn:flmr:late_interaction}
\end{equation}

PreFLMR improves over FLMR in the following aspects: (1) While FLMR only uses the [CLS] embedding from ViT as the image representation, in PreFLMR we additionally extract embeddings of image patches from ViT's penultimate layer to obtain a detailed visual representation. (2) We introduce Transformer blocks with cross-attention into the mapping structure to obtain query-aware visual representation. 
The Transformer blocks take the image patch embeddings as input, and use cross-attention to integrate the features of the text encoder.
This allows PreFLMR to attend to different aspects of the image under different queries. These Transformer blocks are placed in parallel with FLMR's 2-layer MLP mapping structure. 
(3) We append task-specific instructions to the text query to distinguish between tasks. 
The list of instructions for each task can be found in Appendix~\ref{sec:preflmr:appendix:dataset_detail}.
For each query, the instruction is randomly sampled from the corresponding instruction list.
Instruction tokens are masked in computing relevance score. For Q2T retrieval training, we feed a blank image as PreFLMR's image input. For I2T retrieval training, we use instructions as text input to PreFLMR. 

PreFLMR training and inference follow that of FLMR. When training on data consisting of several datasets, we randomly shuffle the entire training data and only use in-batch negative examples from the same corpus. Post-training, all documents are indexed through PLAID~\citep{keshav2022plaid} for efficient late-interaction retrieval. For detailed evaluation of retrieval efficiency, we refer readers to \citet{lin2023finegrained}.

The detailed formal expression of the entire model can be found in Appendix~\ref{sec:preflmr:appendix:model_design}.

\subsection{Training Procedures}
PreFLMR's pre-training involves four stages.

\paragraph{Stage 0: Text Encoder Pre-training.} We train ColBERT following \citet{khattab-zaharia-2020-colbert} on the MSMARCO dataset to obtain the initial checkpoint for PreFLMR's text encoder $\mathcal{F}_L$. 
This is a straightforward replication of ColBERT used as an initial text encoder as was done in FLMR, but also allowing for size variations.


\paragraph{Stage 1: Training the Mapping Structure.} 
In this stage, we only train the mapping structure $\mathcal{F}_M$, keeping the language and vision models frozen. 
This approach is an extension of the FLMR methodology, incorporating a larger dataset and an additional cross-attention mapping layer. The training is performed on the IQ2T dataset (LLaVA, OVEN), I2T datasets (WIT, CC3M, KVQA), and Q2T dataset (MSMARCO). Our objective is to encompass all three task types in M2KR without the need to optimize the data mixing ratio or manually select datasets to achieve an effective mapping structure. This strategy is inspired by previous studies~\cite{lin2023finegrained, zhu2023minigpt, Liu2023llava}, which utilized relatively simple multi-modal tasks to develop image-to-text mappings.

We mask the late-interaction token embeddings in query matrix $\mathbf{Q}$ that are produced by the language model (not the token embeddings at the input embedding layer). This encourages the Transformer cross-attention layer to integrate information from its textual inputs and enables PreFLMR to perform IQ2T, I2T, and Q2T retrieval when provided with the appropriate instructions for each task. 


\paragraph{Stage 2: Intermediate KB-VQA Pre-training.}
We tune the text encoder $\mathcal{F}_L$ and the mapping structure $\mathcal{F}_M$ on the E-VQA dataset, a large and high quality KB-VQA dataset, to enhance PreFLMR's retrieval performance. 
Including an intermediate pre-training stage to align the model with in-domain data has been well-explored in the literature (e.g., Google’s TAPAS~\cite{eisenschlos-etal-2020-understanding}). We opt for a straightforward procedure to train on E-VQA in the intermediate stage because of its diversity, increased difficulty, and larger quantity compared to other KB-VQA datasets. 
Specifically, E-VQA requires recognition of less common entities such as spotted hyenas and relies on more specialized domain knowledge such as American landmarks, making it good for retrieval training. 
This design choice is well-supported by experimental results (Table~\ref{tab:preflmr:main_results} \#8 vs \#5, \#3 vs \#2) and we provide detailed analysis in Sec.~\ref{sec:Analysis of Intermediate Pre-training}.



\paragraph{Stage 3: Full-scale Fine-tuning.}
We train on the entire M2KR corpora, including OKVQA and Infoseek. 
This stage is straightforward multi-task learning.
We tune the entire model except the vision encoder $\mathcal{F}_V$.  We adjust the dataset proportions to ensure balanced learning on these datasets of varying sizes (Appendix~\ref{sec:preflmr:appendix:data_proportion}). Additionally, we use separate text encoders to encode queries and documents; their parameters were shared in previous steps.

\subsection{Training Configurations}
\label{sec:Training Configurations}

We use the Adam optimizer~\cite{KingmaB14adam} with a fixed learning rate of $10^{-4}$ for the mapping structures and $10^{-5}$ for other parameters in all experiments.
Training was run up to 300k, 220k, 12k, and 50k steps in the four stages, respectively. 
Full training configurations (including the hyperparameters for downstream VQA fine-tuning) can be found in Appendix~\ref{sec:preflmr:appendix:hyperparameters}.

\section{Experiments and Results}

In this section we present results of scaling PreFLMR components (Sec.~\ref{sec:preflmr:results:preflmr_performance}, \ref{sec:preflmr:ablation_study}), analyze the effect of each training stage (Sec.~\ref{sec:preflmr:staged_performance}, \ref{sec:preflmr:analysis}), and evaluate on the downstream KB-VQA tasks (Sec.~\ref{sec:preflmr:downstream_kbvqa_performancee}). We summarize our findings in Sec.~\ref{sec:preflmr:summary-of-findings}. Multi-task performance refers to PreFLMR results, i.e. Stages 0, 1, 2, and 3, without any single-task fine-tuning. 

\subsection{Model Variants}
\label{sec:Model Variants}

{We experiment with a range of model configurations. Model sizes range from BERT-Small (28.8M), BERT-Medium (41.1M), BERT-Base (110M) to BERT-large (340M).} ColBERT text encoders are denoted as "[BERT size]-[pre-training scheme]".  There are two ColBERT pre-training schemes: ``v1''~\cite{khattab-zaharia-2020-colbert} and ``v2''~\cite{santhanam-etal-2022-colbertv2}. ``v2'' yields a better performing model than ``v1'' as evaluated on MSMARCO. 
We compare models initialized from ``v1'' and ``v2'' checkpoints to investigate how the performance of the initial uni-modal text retriever affects the final multi-modal vision-language retriever. Except for ``Base-v2'', all ColBERT variants are trained using our replication of ColBERT following the ``v1'' pre-training scheme.\footnote{The training code of ``v2'' has not been released officially.}
{For the vision encoders, we use the ViT variants: ViT-B (88M), ViT-L (303M)~\cite{Radford_2021_CLIP}, ViT-H (631M) and ViT-G (1.84B)~\cite{cherti2023openclip}.} 

\subsection{PreFLMR Performance}
\label{sec:preflmr:results:preflmr_performance}


\begin{table*}[!ht]
\centering
\resizebox{1.0\linewidth}{!}{%
\begin{tabular}{@{}cllllcccccccccr@{}}
\toprule 
\multicolumn{5}{c}{} & \multicolumn{3}{|c}{I2T} & \multicolumn{1}{|c|}{Q2T} & \multicolumn{5}{c|}{IQ2T} & \\
\midrule 
 & \multirow{2}{*}{Model} & Vis.   & Text & Total & WIT & IGLUE  & KVQA & \multicolumn{1}{c}{MM} & OVEN & LLaVA & Infoseek & E-VQA & OKVQA & A.R. \\
 & &Enc. & Enc. & Param. &  R@10 & R@1 & R@5 & R@5 & R@5 & R@1 & PR@5 & PR@5 & PR@5 \\
\midrule
 & CLIP & & & & 28.1 & 44.1 &  23.8 & - & 22.0 & 33.0  & 17.1   & 10.4 & 5.7 &  \\
 & SOTA & & & & FLMR & GIVL &  FLMR & ColBERT & FLMR & FLMR & FLMR & Lens & FLMR & \\
 & ~~~ Res. & & & & 23.8 & 30.8 & 31.9 & 86.9 & 40.5 & 56.4 & 47.1 & 62.5\tablefootnote{The performance is not fully comparable due to differences in the construction of the test passage corpus and the proprietary nature of the data and pipeline used in Lens. The reported figures serve as a reference point.} & 68.1 &  \\
\midrule
\multicolumn{13}{l}{\textit{Multi-task Performance}} \\
\midrule
1 & PreFLMR & B & B-v1 & 207M & 41.5 & 56.8  & 28.6 & 77.9 & 45.9 & 67.4 & 48.9 & 65.4 & 67.2 & 9.0 \\
2 & PreFLMR & B & B-v2 & 207M & 41.7 & 57.3 &  28.6 & 79.5 & 46.3 & 67.2 & 48.8 & 67.9 & 66.1 & 8.2 \\
3 & ~\textit{w/o inter.} & B & B-v2 & 207M &  41.2 & 56.8 & 26.5 & 78.2 & 43.7 & 65.0 & 47.0 & 57.3 & 65.1 & 10.9 \\
4 & PreFLMR & L & B-v1 & 422M & 58.2 & 69.8 & 40.6 & 72.1 & 59.3 & 69.3 & 57.4 & 70.7 & 67.9 & 5.6 \\
5 & PreFLMR & L & B-v2 & 422M & 60.5 & 69.2 &  \textbf{43.6} & 78.7 & 59.8 & 71.8 & 57.9 & 70.8 & 68.5 & 3.2\\
6 & ~\textit{ViT trainable} & L & B-v2 & 422M & 18.7 & 1.5 &  0.8 & 76.7 & 5.6 & 54.6 &  36.7 & 57.2 & 58.9 & 12.3 \\
7 & ~\textit{w/o instruct.} & L & B-v2 & 422M & 13.3 & 10.5 & 38.2 & 75.2 & 52.1 & 62.1 & 49.1 & 71.3 & 65.7 & 9.2 \\
8 & ~\textit{w/o inter.} & L & B-v2 & 422M & 60.0 & 72.0  & 40.5 & \textbf{80.3} & 56.1 &  70.5 & 55.4 & 67.0 & 66.6 & 4.6 \\
9 & PreFLMR & L & S-v1 & 334M & 54.2 & 66.3 &  37.9 & 73.6 & 53.9 & 66.0 &  52.6 & 66.8 & 65.3 & 8.3 \\
10 & PreFLMR & L & M-v1 & 348M & 56.2 & 67.9 &  37.1 & 72.9 & 55.5 & 64.7 & 52.2 & 70.4 & 65.3 & 8.2\\
11 & PreFLMR & L & L-v1 & 677M & 49.9 & 62.8 & 40.0 & 72.8 & 58.8 & 69.3 & 59.4 & 58.2 & 68.6 & 6.6 \\
12 & PreFLMR & H & B-v2 & 750M & 60.5 & 71.2 &  39.4 & 78.5 & 61.5 & 72.3 & \textbf{59.5} & 71.7 & 68.1 & 3.1 \\
13 & PreFLMR & G & B-v2 & 1.96B & \textbf{61.5} & \textbf{71.5} & 42.1 & 78.6 & \textbf{63.4} & \textbf{72.4} & \textbf{59.6} & \textbf{73.1} & \textbf{68.6} & \textbf{1.6} \\
\midrule
\multicolumn{13}{l}{\textit{{Fine-tuned PreFLMR for Specific Downstream Tasks}}} \\
\midrule
14 & PreFLMR & L & B-v2 & 422M & 68.5 & &  & & 70.8 & & 60.3 & 71.4 & 67.3 \\
15 & PreFLMR & H & B-v2 & 750M &69.3 & &  & & 72.3 & & 62.3 & 72.1 & 70.5 \\
16 & PreFLMR & G & B-v2 & 1.96B & \underline{69.3} & &  & & \underline{73.1} & & \underline{62.1} & \underline{73.7} & \underline{70.9} \\
\bottomrule
\end{tabular}
}
\caption{PreFLMR performance on all datasets. PR stands for Pseudo Recall. Best multi-task performance is in bold and best downstream fine-tuning performance is underlined. {For the vision encoder, we compare ViT-B (B), ViT-L (L), ViT-H (H) and ViT-G (G). For the text encoder, we compare Base-v1 (B-v1), Base-v2 (B-v2), Small-v1 (S-v1), Medium-v1 (M-v1), and Large-v1 (L-v1).} A.R.: Average Rank against all other models on all tasks. For baselines, we show: GIVL \cite{Yin2023GIVL} for IGLUE; ColBERTv2 for MSMARCO (MM); FLMR~\cite{lin2023finegrained} for Infoseek and OKVQA; and Google Lens~\cite{googlelens} for E-VQA. We follow the procedure as detailed in the Appendix C of the E-VQA paper~\cite{Mensink2023EVQA} to use CLIP as a zero-shot retriever.}
\label{tab:preflmr:main_results}
\vspace{-0.2cm}
\end{table*}

The best-performing PreFLMR model (ViT-G + Base-v2) outperforms other variants on most of M2KR benchmark  (Table \ref{tab:preflmr:main_results}, \#13). 
Without single-task fine-tuning, PreFLMR outperforms baseline models optimized for the individual tasks on 7 out of 9 datasets, showcasing its capability as a general visual-language retriever.  We now analyze how each PreFLMR component affects performance. 

\paragraph{Vision Encoder Scaling.} Scaling ViT from ViT-B (86M) to ViT-G (1.8B) while keeping the text encoder fixed brings about substantial performance gain across all tasks (Table~\ref{tab:preflmr:main_results} \#2, \#5, \#12, \#13), e.g. 48.8 to 59.6 on Infoseek and 67.9 to 73.1 on E-VQA. 
The gain is greater when upgrading ViT-B to ViT-L with recall improvements of $\sim 10\%$ on WIT, KVQA, OVEN, and Infoseek, showing the benefit of using better vision encoders.
In addition, Fig.~\ref{fig:preflmr:radar_plot} in the appendix illustrates performance gains in scaling the vision encoder with a radar plot.
However, the performance plateaus when scaling ViT to H and G. This observation aligns with results reported in the literature. OpenCLIP~\cite{cherti2023openclip} and BLIP2~\cite{li2023blip} have reported marginal or no performance improvement when scaling beyond ViT-L across several datasets. A plausible explanation is that if the ViT model is not pre-trained on domain-specific data, it may struggle to make fine distinctions.


\paragraph{Text Encoder Scaling.} 
Scaling up the text encoder from BERT-Small-v1 to Medium-v1 to Base-v1 (Table~\ref{tab:preflmr:main_results} \#9, \#10, \#4) yields substantial performance gain (A.R. 8.3, 8.2, and 5.6). However, we find that further scaling to Large-v1 (\#11) adversely impacts the performance (A.R. decreased to 6.6). We attribute this to overfitting and unstable training for large models given the available data (Appendix~\ref{sec:preflmr:appendix:large_v1_training}). The results suggest that BERT-Base (110M) is adequate for building a capable vision-language retriever. 


\paragraph{Improving Text Encoder.} Compared to PreFLMR models initialized from Base-v1, models initialized from Base-v2 have better multi-tasking performance indicated by better A.R. (Table~\ref{tab:preflmr:main_results} \#1 vs \#2 and \#4 vs \#5). The gain from improving the text encoder is more substantial when using the ``ViT-L'' vision model (-2.4 A.R.) compared to using ``ViT-B'' (-0.8 A.R.), 
indicating that the text encoder is relatively weak as the vision model improves. 




\subsection{Performance of Each PreFLMR Stage}
\label{sec:preflmr:staged_performance}

In this section, we analyze intermediate performance in the earlier stages of pre-training to better understand the scaling behaviour of PreFLMR.

\begin{figure*}[htp]
\centering
\begin{minipage}{0.30\textwidth}
\centering
\resizebox{\linewidth}{!}{%
\begin{tabular}{@{}lrr@{}}
\toprule
Model              & MRR@10 & Recall@50 \\ \midrule
Small-v1 (28.8M)   & 34.5   & 79.8      \\
Medium-v1 (41.4M)  & 35.5   & 81.4      \\
Base-v1 (110M)     & 35.8   & 82.4      \\
Large-v1 (345M)   & 37.0     & 83.2      \\ \midrule
Base-v1 (official) & 36.0     & 82.9      \\
Base-v2 (official) & 39.7   & 86.8      \\ \bottomrule
\end{tabular}
}
\captionof{table}{Text encoder pre-training results evaluated on the full MSMARCO test set.}
\label{tab:preflmr:colbert_results} 
\end{minipage}
\hfill
\begin{minipage}{0.68\textwidth}
\centering
\resizebox{\linewidth}{!}{%
\begin{tabular}{@{}lllrrrrrrrrr@{}}
\toprule
& Vis. Enc.   & Text Enc. & WIT   & LLa. & OVEN  & KVQA  & IGLUE & Info. & E-VQA  & OK. & A.R. \\ \midrule
1 &  ViT-B & Base-v2   & 34.2          & 50.9          & 46.1          & 28.9          & 60.5          & 42.5          & 32.7          & 46.5          & 6.5          \\
2  & ViT-L & Small-v1  & 46.5          & 46.1          & 37.9          & 17.9          & 57.3          & 43.5          & 26.6          & 56.7          & 7.0          \\
3  & ViT-L & Medium-v1 & 49.6          & 47.8          & 38.6          & 23.1          & 58.7          & 46.7          & 27.7          & \textbf{58.1} & 5.3          \\
4  & ViT-L & Base-v1   & 49.3          & 50.8          & 52.3          & 38.2          & 68.5          & 46.1          & 41.9          & 49.4          & 4.6          \\
5  & ViT-L & Base-v2   & 49.6          & 51.2          & 54.8          & \textbf{40.5} & \textbf{69.5} & 48.7          & \textbf{45.0} & 50.9          & 2.3          \\
6  & ViT-L & Large-v1  & 48.5          & 47.3          & 51.8          & 32.8          & 67.2          & 45.1          & 40.0          & 49.7          & 5.6          \\
7  & ViT-H & Base-v2   & \textbf{51.8} & 51.6          & 55.3          & 35.6          & 69.0          & 48.6          & 42.2          & 51.3          & 2.8          \\
8  & ViT-G & Base-v2   & 49.5          & \textbf{51.8} & \textbf{59.6} & 38.7          & 69.3          & \textbf{50.9} & 42.4          & 52.1          & \textbf{2.0}
 \\ \bottomrule
\end{tabular}
}
\captionof{table}{PreFLMR performance after Stage 1. Infoseek, E-VQA, and OKVQA are tested in zero-shot mode. A.R.: Average Rank against all other models on all tasks. LLa.- LLaVA; Info.- Infoseek; OK. - OKVQA.}
\label{tab:preflmr:stage1_results}
\end{minipage}

\end{figure*}

\paragraph{Text Encoder Pre-training.} We train ``ColBERT-v1'' at different sizes and evaluate on the MSMARCO dataset. Table \ref{tab:preflmr:colbert_results} shows larger model sizes consistently yield better text retrieval performance. In contrast to the multi-modal case, scaling up to ``Large-v1'' does not destabilize training and leads to better performance compared to ``Base-v1''.


\paragraph{Training the Mapping Structures.} 
Table~\ref{tab:preflmr:stage1_results} details system performance after Stage 1 training, in which only the vision-language mapping structure is trained. 
Similar to Sec.\ref{sec:preflmr:results:preflmr_performance}, scaling up the vision encoder improves performance across tasks. 
PreFLMR exhibits strong zero-shot KB-VQA performance at this preliminary stage (50.87 in Infoseek, 42.44 in E-VQA, and 52.14 in OKVQA). 
After Stage 1, PreFLMR with ViT-G performs worse than other variants on IGLUE, E-VQA and OKVQA. However, it attains the best performance on these datasets after Stage 3. This suggests that tuning the mapping structure alone is not enough to fully utilize larger vision models.



\paragraph{Intermediate Pre-training.} 
Stage 1 improves performance on KB-VQA tasks (Table~\ref{tab:preflmr:main_results} \#3 vs \#2 and \#8 vs \#5). Although we are only training on E-VQA, the score on other KB-VQA tasks (Infoseek, KVQA, OKVQA) increases by $\sim$1\% or more. This shows that E-VQA is an appropriate corpus for training a general-purpose knowledge retriever. We analyze the gain from intermediate pre-training in more detail in Sec.\ref{sec:preflmr:analysis}.

\subsection{Ablation Studies}
\label{sec:preflmr:ablation_study}


\paragraph{Instructions.}
Removing instructions (Table~\ref{tab:preflmr:main_results} \#7) results in much worse overall performance, with the  WIT recall rate reduced to 13.3. This shows that instructions are necessary for multi-task learning and that our instruction scheme works well (the full list of instructions  is given in Appendix~\ref{sec:preflmr:appendix:dataset_detail}). 


\begin{table}[!h]
\centering
\resizebox{0.75\linewidth}{!}{%
\begin{tabular}{@{}llll@{}}
\toprule
Datasets       & WIT   & LLaVA & Infoseek \\ \midrule
All            & 34.14 & 50.82 & 42.71    \\
~~~\textit{w/o CC3M}         & 29.33 & 44.82 & 40.18    \\
~~~\textit{w/o LLaVA}      & 33.78 & 30.78 & 39.20     \\
~~~\textit{w/o MSMARCO}         & 33.96 & 47.88 & 38.90     \\
~~~\textit{w/o OVEN\&KVQA} & 33.96 & 49.85 & 35.62    \\ \bottomrule
\end{tabular}
}
\caption{Ablation study on Stage 1 pre-training datasets. The model is ViT-B + Base-v1. {We evaluate systems on Infoseek in zero-shot mode though it is not used in Stage 1 training.}}
\label{tab:preflmr:dataset_ablation}
\end{table}

\paragraph{Pre-training Datasets.}
As shown in Table~\ref{tab:preflmr:dataset_ablation}, adding CC3M to training improves performance on all metrics, showing that learning to understand scene via captioning datasets is beneficial.
Removing either LLaVA or MSMARCO harms zero-shot KB-VQA performance ($-3.0$ in Infoseek), {noting that Infoseek is not used in this stage.} Training on these datasets facilitates learning question-aware visual representations as the cross-attention in the mapping structure must attend to the text input to perform well on these tasks. 
Omitting knowledge-intensive datasets (OVEN and KVQA) negatively impacts the zero-shot performance on Infoseek, showing the importance of in-domain data. 

\paragraph{Mapping Structure Scaling.}

\begin{table}[!h]
\centering
\resizebox{0.85\linewidth}{!}{%
\begin{tabular}{@{}llccc@{}}
\toprule
                & $N_{TR}$  & WIT  & LLaVA & Infoseek \\ \midrule
ViT-B + Base-v1 & 1L & 34.1 & 50.8  & 42.7     \\
ViT-B + Base-v1 & 4L & 29.0 & 51.4  & 40.8     \\
ViT-L + Base-v2 & 1L & 49.6 & 51.2  & 48.7     \\
ViT-L + Base-v2 & 4L & 45.9 & 51.7  & 46.8     \\ \bottomrule
\end{tabular}
}
\caption{Performance of adding more Transformer layers to the mapping structure. $N_{TR}$ is the number of Transformer layers in the mapping structure.}
\label{tab:preflmr:mapping_ablation}
\end{table}

Table~\ref{tab:preflmr:mapping_ablation} illustrates the impact of scaling up the mapping structure under two PreFLMR configurations.
Increasing cross-attention layers from 1 to 4 marginally improves LLaVA performance (+0.5, approx.), but adversely impacts performance on WIT (-4, approx.) and Infoseek (-2, approx.). 
We adhere to the 1-layer design, noting that adding parameters to the mapping structure does not improve performance.

\subsection{{Retrieval Augmented Visual Question Answering with PreFLMR}}
\label{sec:preflmr:downstream_kbvqa_performancee}

\begin{table}[!h]
\resizebox{\linewidth}{!}{%
\begin{tabular}{@{}lccc@{}}
\toprule
Model                       & OKVQA  & Infoseek & E-VQA        \\ \midrule
Baseline                        & 66.10  & 21.80    & 48.80       \\
~~~\textit{Baseline model}                  & PaLM-E & PALI-X   & PaLM-B + Lens \\
AVIS & 60.20 & 50.70/56.40\tablefootnote{50.7 for Unseen Entity and 56.4 for Unseen Question; no overall accuracy is reported.} & - \\
RA-VQAv2 w/ FLMR    & 60.75  & -        & -           \\
RA-VQAv2 w/ PreFLMR & 61.88  & 30.65    & 54.45       \\
~~~\textit{w/o retrieval}               & 55.44  & 21.78    & 19.80       \\ \bottomrule
\end{tabular}
}
\caption{Downstream KB-VQA performance when RA-VQAv2~\cite{lin2023finegrained} is equipped with PreFLMR and fine-tuned on the target M2KR's KB-VQA sub-tasks.
AVIS~\cite{hu2024avis} is a recently published hybrid system that leverages many planning stages to solve KB-VQA questions, which we include for reference.
Performance on Infoseek and E-VQA may not be directly comparable to results in the literature.$^\text{5}$
}
\label{tab:preflmr:downstream_vqa_results}
\end{table}

We build on RA-VQAv2~\cite{lin2023finegrained}, a strong retrieval-augmented visual question answering system to tackle OKVQA, Infoseek, and E-VQA. 
We fine-tune the best-performing PreFLMR variant on the target retrieval task (ViT-G + Base-v2, Table \ref{tab:preflmr:main_results} \#14) 
and follow RA-VQAv2 to fine-tune a BLIP-2 answer generator on the target M2KR KB-VQA task.\footnote{We note that this work was conducted during the early stage of the release of Infoseek and E-VQA. We prepared the data splits according to the need for retrieval training following Appendix~\ref{sec:preflmr:appendix:dataset_detail}.
The systems are trained and evaluated on the data splits provided in M2KR to show the improvement relative to systems without retrieval.
}
Following previous literature~\citep{schwenk2022okvqa, Chen2023infoseek, Mensink2023EVQA}, we use VQA score, Accuracy, and BERT matching (BEM)~\cite{bulian-etal-2022-tomayto} to evaluate performance on OKVQA, Infoseek, and E-VQA, respectively.


A brief summary of the systems shown in Table~\ref{tab:preflmr:downstream_vqa_results}:
PaLM-E~\citep{driess2023palme}, PALI-X~\cite{Chen2022PaLI} and PaLM-B~\cite{anil2023palm} are large multi-modal models with 562B, 55B, and 1T parameters, respectively. The E-VQA SOTA~\citep{Mensink2023EVQA} uses Lens~\citep{googlelens}, the Google API for image retrieval.
AVIS~\cite{hu2024avis} is a hybrid system with many components (such as PaLI, PaLM, and Google Lens\&Web Search
API) and planning stages powered by LLMs. 
We note that PreFLMR could be used as part of the AVIS pipeline to enhance its ability to fetch relevant documents given questions and images.


As shown in Table~\ref{tab:preflmr:downstream_vqa_results}, compared to models without retrieval, PreFLMR improves performance by approximately 6\% on OKVQA, 9\% on Infoseek, and 34\% on E-VQA. These results highlight the effectiveness of PreFLMR in document retrieval for KB-VQA tasks.


On OKVQA, the performances of RA-VQAv2 (PreFLMR) and RA-VQAv2 (FLMR) are similar. Table~\ref{tab:preflmr:main_results} \#13 shows that PreFLMR attains similar Recall@5 as FLMR on OKVQA even though it has a much larger vision encoder. As a possible explanation, compared to E-VQA and Infoseek, the knowledge required to answer OKVQA question is less specialized and many OKVQA questions can be answered without document retrieval~\cite{Mensink2023EVQA}.
See Appendix \ref{sec:preflmr:comparing_okvqa_evqa} for qualitative analysis.
Another possibility is that, compared to E-VQA and Infoseek where the ground-truth document is provided for each question, the OKVQA training set does not provide ground-truth knowledge documents. The retriever uses pseudo-relevant documents in training that contain the target answer but these may not be truly useful for answering the question. 
This is evidence that data quality should be improved along with model scaling.





\subsection{Analysis of Intermediate Pre-training}
\label{sec:preflmr:analysis}
\label{sec:Analysis of Intermediate Pre-training}

Sec.~\ref{sec:preflmr:staged_performance} shows that Stage 2 Intermediate Pre-training improves the performance as evaluated by task-specific metrics. 
In this section, we further quantify the gains from Stage 2  for each dataset and more clearly show that KB-VQA tasks benefit more from Stage 2 than other tasks. We use the difference in minimal validation loss\footnote{We find that the validation loss is predictive of the actual performance. A lower validation loss usually suggests a better performance in the tasks that we study.} achieved on each dataset starting from checkpoints before or after Stage 2 Intermediate Pre-training as a measure of benefit. This enables comparison of tasks with different performance metrics. Intuitively, a larger absolute difference in validation loss indicates that the dataset benefits more from the Intermediate Pre-training stage.


\begin{figure}[!h]
    \centering
    \includegraphics[width=1.05\linewidth]{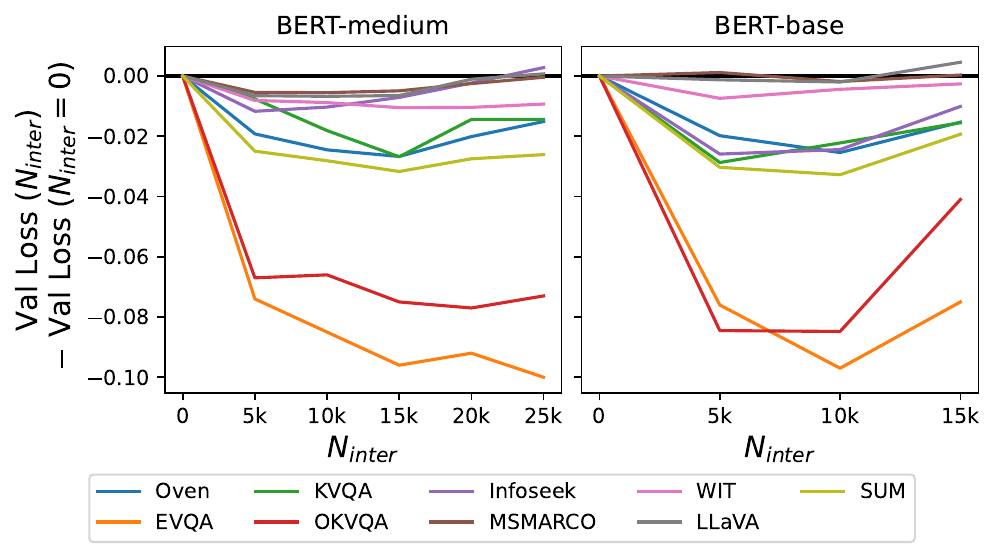}
    \caption{Change in Stage 3 validation loss when initialized from Stage 2 checkpoints after $N_{inter}$ steps of intermediate pre-training. 
    A large difference indicates a greater gain from intermediate pre-training.
    }
    \label{fig:preflmr:intermediate_plot}
\end{figure}

Figure \ref{fig:preflmr:intermediate_plot} plots the difference in validation loss of every dataset when the starting checkpoints have undergone $N_{inter}$ intermediate pre-training steps using either BERT-medium or BERT-base as the text encoder backbone. 
As expected, starting from E-VQA-pre-trained checkpoints yields lower validation loss in knowledge-intensive tasks such as OKVQA, KVQA, and OVEN after the same number (5,000) of fine-tuning steps. Performance on these datasets indeed sees more gain from Stage 2 training (Table \ref{tab:preflmr:main_results}, \#5 v.s. \#8). Figure \ref{fig:preflmr:intermediate_plot} also indicates the existence of an optimal $N_{inter}$, beyond which the model overfits to E-VQA, harming performance on other datasets. The larger PreFLMR model with BERT-base text encoder overfits faster than PreFLMR with BERT-medium ($N_{inter}\approx 15,000$ versus $N_{inter}\approx 10,000$). We use V-Entropy~\cite{VInformation} to formalize our analysis as an empirical measure of mutual information between datasets in Appendix~\ref{sec:preflmr:appendix:v_information}.

\subsection{Summary of Findings}
\label{sec:preflmr:summary-of-findings}

We summarise the results of our investigations into scaling behaviour as follows:
\begin{itemize}[leftmargin=5.5mm]
\setlength\itemsep{-0.3em}
\item The text encoder size need not exceed that of BERT-base (110M) to achieve {competitive multi-modal retrieval performance} (Sec.\ref{sec:preflmr:results:preflmr_performance}).


\item  Scaling up the vision encoder from ViT-B to ViT-G yields substantial gains (Sec.\ref{sec:preflmr:results:preflmr_performance}).

\item Scaling up the mapping structure does not improve performance (Sec.\ref{sec:preflmr:ablation_study}). 

\item Intermediate pre-training on high-quality in-domain data (E-VQA) effectively improves retrieval performance across KB-VQA tasks (Sec.\ref{sec:preflmr:staged_performance}, \ref{sec:Analysis of Intermediate Pre-training}). 

\item Strong knowledge retrievers boost performance on challenging KB-VQA tasks such as OKVQA, Infoseek, and E-VQA via Retrieval-Augmented Generation (Sec.\ref{sec:preflmr:downstream_kbvqa_performancee}).

\item Ground-truth document labels are {important} to make full use of large models in training multi-modal retrievers (Sec.\ref{sec:preflmr:downstream_kbvqa_performancee}).
\end{itemize}

\section{Conclusion}
This work has studied the scaling behaviour of state of the art multi-modal document retrieval systems, with a focus on enhancing fine-grained late-interaction retrieval for knowledge-based visual question answering.  We contribute a comprehensive training and evaluation framework, M2KR, for general-purpose multi-modal knowledge retrieval.  The PreFLMR system we train in the M2KR framework yields excellent retrieval performance across a range of tasks and can also serve as a base for further task-specific fine-tuning.

\newpage

\section*{Limitations}
Limited by available computational resources, we leave several further investigations as future work:
(1) The CLIP-ViT models~\cite{cherti2023openclip} were not pre-trained on in-domain data of knowledge-intensive tasks. Further training may enhance the model's ability to recognize a broader range of objects;
(2) Advanced training approaches beyond contrastive learning, such as score distillation~\cite{santhanam-etal-2022-colbertv2}, could be explored to further enhance retrieval performance;
(3) Investigating a more optimal mix proportion of datasets with varying sizes also warrants further exploration.

\section*{Ethics Statement}
Our proposed model retrieves documents without generating new content.
We acknowledge the potential for the retrieved documents to include inappropriate information if the document database lacks adequate filtering. 
Consequently, extra care must be taken to ensure the sanitization of the document database, particularly when employing this model in applications involving direct interaction with real users.

\section*{Acknowledgments}
This work was supported in part by the AWS Cloud Credit for Research programme.

Weizhe Lin is supported by a Research Studentship funded by Toyota Motor Europe (RG92562(24020)) for the undertaking of the PhD in Engineering at the University of Cambridge.

Jingbiao Mei is supported by Cambridge Commonwealth, European and International Trust for the undertaking of the PhD in Engineering at the University of Cambridge.

Jinghong Chen is supported by the Warwick Postgraduate Studentship from Christ’s College and the Huawei Hisilicon Studentship for the undertaking of the PhD in Engineering at the University of Cambridge.

Prof. Bill Byrne holds concurrent appointments as a Professor of Information Engineering at Cambridge University and as an Amazon Scholar.  This publication describes work performed at Cambridge University and is not associated with Amazon.

We would also like to thank all the reviewers for their knowledgeable reviews.
\bibliography{custom}

\appendix

\section{Datasets details}
\label{sec:preflmr:appendix:dataset_detail}
This section outlines the preprocessing methods used to convert various datasets into formats suitable for retrieval tasks. Table~\ref{tab:dataset_demo} provides examples from each dataset, demonstrating the transformation from their original to the adapted structure. Subsequent subsections detail the specific preprocessing steps for each dataset. The M2KR dataset is available at \href{https://huggingface.co/datasets/BByrneLab/multi\_task\_multi\_modal\_knowledge\_retrieval\_benchmark\_M2KR}{Huggingface Hub}.
\subsection{I2T Retrieval}
\subsubsection{WIT}

WIT \cite{srinivasan2021wit} is a corpus based on Wikipedia with image-text pairs, where the text is the Wikipedia passage associated with the image.
To enhance data quality, we exclusively select image-text pairs where the images are the main/title images of their respective Wikipedia documents, and we limit our scope to English-language documents.

Our training set, comprising 2.8 million examples, is sourced from the original WIT training set. 20,102 and 5,120 examples from the original WIT validation set are selected to build the validation set and test set in our M2KR benchmark, respectively. 
The test corpus includes all documents from the original WIT validation and test sets. This setting ensures that there is no overlap between different sets.

Each image-document pair is paired with a randomly selected instruction from our set of templates. The task is to retrieve the correct document from the test corpus, given the image and instruction.


\subsubsection{IGLUE} 
The IGLUE English retrieval test set \cite{Bugliarello2022iglue}, which is a subset of the WIT test set and has an established benchmark for image-to-text retrieval, is included to enable comparison with models in previous literature. 
Following \citet{Bugliarello2022iglue}, the test set contains 685 unique images and 1,000 Wikipedia passages.
The task is similar to WIT: using the image and the instruction to retrieve the corresponding Wikipedia passage.
\\
\textbf{\textit{Instruction templates for WIT and IGLUE}}:

• <Image> Identify the document that is connected to this image.

• <Image> Provide information about the document linked to this image.

• <Image> Please describe the document that corresponds to this image.

• <Image> What is the document that this image is related to?

• <Image> Could you elucidate the document associated with this image?

• <Image> Describe the document that accompanies this image.

• <Image> Please give information on the document that goes with this image.

• <Image> What document is represented by this image?

• <Image> Identify the document that this image pertains to.

\subsubsection{KVQA}
\label{sec:preflmr:dataset_detail:kvqa}
KVQA \cite{Shah2019KVQA} is a dataset containing a rich collection of entities representing famous individuals. 
The KVQA task, initially designed as a KB-VQA task, has been re-purposed into an I2T task for our modelling purposes. This adaptation is based on our findings that using images as queries alone suffices to retrieve the documents containing the correct identities. In our context, where the primary focus is on document retrieval, the original questions are unnecessary.
Our reformulated task for KVQA is to retrieve the details of famous people like gender, nationality, birthplace, and employment history based solely on their images. The training set is downsampled from the KVQA original training set by removing repeated examples of the same famous individuals. We transformed the structured entities such as gender and nationality into passages.
For example, ``nationality: America; date of birth: dd/mm/yyyy; ...'' is serialized as ``nationality is America, date of birth is dd/mm/yyyy, ...''.

The training corpus is composed of all the documents that appear in the original KVQA training set. For the validation/test set, we selected a subset of 13,365/5,120 samples from the original KVQA validation set. Correspondingly, the test corpus encompasses all documents found in the original KVQA validation set.

The instruction we use for KVQA is:
<Image> Provide a brief description of the image and the relevant details of the person in the image.

\subsubsection{CC3M}

CC3M \cite{Sharma2018CC3M} is a dataset consisting of a vast collection of image-caption pairs. Instead of utilizing the entire dataset comprising 3 million pairs, we adopt the downsampling methodology as delineated in LLaVA's work \cite{Liu2023llava}, resulting in a reduced dataset of approximately 595K. 

We reformulate the image-caption pairs into image-to-text retrieval tasks in our pre-training. To construct the training corpus, we treat each caption as an individual document linked to its corresponding image. The task then involves retrieving the most relevant caption for a given image, guided by a set of randomly selected instructions. Since CC3M is originally an image captioning task, we do not validate or test our retriever on CC3M. 
\\
\textbf{\textit{Instruction templates for CC3M }}

• <Image> Describe the image concisely.

• <Image> Provide a brief description of the given image.

• <Image> Offer a succinct explanation of the picture presented.

• <Image> Summarize the visual content of the image.

• <Image> Give a short and clear explanation of the subsequent image.

• <Image> Share a concise interpretation of the image provided.

• <Image> Present a compact description of the photo’s key features.

• <Image> Relay a brief, clear account of the picture shown. 

• <Image> Render a clear and concise summary of the photo.

• <Image> Write a terse but informative summary of the picture. 

• <Image> Create a compact narrative representing the image presented.

\subsection{Q2T Retrieval}
\subsubsection{MSMARCO}

MSMARCO \cite{Bajaj_2018MSMARCO} stands for Microsoft Machine Reading Comprehension dataset. It is a text-only dataset with around 1 million questions and 8 million passages. At stage 0, we train according to ColBERT-v1 by \citet{khattab-zaharia-2020-colbert}. For later stages, we downsample the dataset to 400K questions to balance between the multimodal tasks and unimodal tasks. For the training corpus, we still use the full 8 million passages.
For testing, we select 6,980 and 5,120 samples from the original MSMARCO validation set and sample 400K passages to retrieve from and ensure the subset contains all ground-truth passages. 
\\
\textbf{\textit{Instruction templates for MSMARCO}}:

• <Blank image> Retrieve the document that answers this question. <Questions>

• <Blank image> Find the document that is most relevant to the question. <Questions>

• <Blank image> Obtain the document that resolves this query. <Questions>

• <Blank image> Acquire the document that elucidates this question. <Questions>

• <Blank image> Choose the document most relevant to the query. <Questions>

• <Blank image> Identify the document most applicable to the question. <Questions>

• <Blank image> Extract the document that answers this query. <Questions>

• <Blank image> Locate the document that addresses the query.<Questions>

\subsection{IQ2T Retrieval}
\subsubsection{LLaVA}
\label{sec:preflmr:dataset_detail:llava}
The LLaVA instruction following dataset contains GPT-3.5 generated high-quality conversation about an image between a human and an AI assistant. There are around 150K rounds of conversations. We took each conversation (each question from the human and the answer from the AI assistant) as a separate sample. This results in a total of 356K samples. Since there are no original validation or test sets associated with the LLaVA, we manually split the sample pool into 351K training examples and 5,120 test examples. 

The task is reformulated to an Image\&Question to Text retrieval task. The training corpus and test corpus each contain the associated answers as passages to be retrieved by the image and question pairs. We use two types of instruction templates depending on the preciseness of the question: 

• <Image> Provide a brief description of the image along with the following question: <Question>

• <Image> Provide a concise explanation of the image along with the following question: <Question>


\subsubsection{OVEN} 

OVEN is a dataset targeting open-domain visual entity recognition. The dataset consists of two splits: entity set and query set. The entity set is derived from image classification datasets such as INaturalist2017 \cite{Cui2018inaturalist}, Food-101 \cite{bossard14food101}, Cars196 \cite{KrauseStarkDengFei-Fei2013cars196} and Google Landmarks Dataset v2 \cite{Weyand2020googlelandmarkv2}. The query set is derived from VQA datasets such as VQAv2 \cite{balanced_vqa_v2} and OKVQA \cite{schwenk2022okvqa}. To avoid overlapping with our other KB-VQA datasets, we only use the entity set of OVEN.   The entity set contains about 10K unique entities. 

The original entity set contains about 5 million question-image pairs. However, the questions are highly duplicated in the original OVEN dataset. We downsample the dataset by removing repeated questions corresponding to the same entity. This reduces duplications while maximizing the diversity of the questions and coverage of entities. 
After the filtering, we keep 339K training samples.
For validation and testing, we select 20,000 and 5,120 examples from the original OVEN Entity validation set. The original test set is not used in M2KR due to the lack of annotation. 

The original task is to link the image to a specific Wikipedia Entity given a question.
To formulate the task as a retrieval problem, 
for each entity, we use its associated Wikipedia passage as the document to retrieve.
The query side of this retrieval task contains the image and its question with the inclusion of a randomly sampled instruction. Given this query, the task is to obtain the relevant Wikipedia passage. 
The training corpus contains about 10K passages, while the test corpus contains about 3.2K passages that cover all entities in OVEN's original training set and validation set respectively.

\subsubsection{E-VQA, Infoseek and OKVQA}

E-VQA, Infoseek, and OKVQA are Knowledge-based VQA (KB-VQA) datasets. For each given image and question (with instruction), the task is to retrieve the corresponding knowledge passage. 

For E-VQA \cite{Mensink_2023_ICCV}, the original training set contains around 1 million samples. However, it includes duplicated questions and answers referring to the same Wikipedia Entity with different query images. We filter duplicated questions that pertain to the same Wikipedia Entity. To align with the original evaluation setting of E-VQA, we further excluded samples that necessitate multiple knowledge bases, reducing the count to 167K training samples. To be consistent with the original E-VQA paper, our validation and testing sets exclusively include questions that can be answered using single knowledge. These sets contain 9,852 and 3,750 samples, respectively. We use the WikiWeb2M \cite{Burns2023WIKIWEB2M} as the knowledge source. For the training and test passage corpus, we keep all the passages that appear in the original E-VQA to align with the official E-VQA's setting for retrieval.

For \textbf{OKVQA}, we use the original training and test set. Following~\citet{lin2023finegrained}, we prepare a knowledge corpus with Wikipedia documents based on pseudo-relevance. 
The training and test passage corpus both contain all passages in the knowledge corpus.

For \textbf{Infoseek}, following the preprocessing steps described by~\citealt{Chen2023infoseek}, we use Wikipedia documents as knowledge sources and remove examples whose answers can not be found in the ground-truth documents.
We randomly selected 100K examples from the training set for training and 4,708 examples from the validation set for testing (the annotation of the original test set has yet been released).
The downsampling is motivated by our observation that many questions are repeated and the number of unique documents associated with the whole dataset is only about 40K.
We downsampled the dataset such that the model won't overfit severely to Infoseek passages.


Note that the aforementioned downsampling procedure for the test set is only used for constructing the M2KR benchmark. For downstream VQA evaluation, we use the same test set that existed in previous literature to ensure a fair comparison.
\\
\textbf{\textit{Instruction templates for OVEN, Infoseek, E-VQA, and OKVQA}}

• <Image> Using the provided image, obtain documents that address the subsequent question: <Question>

• <Image> Retrieve documents that provide an answer to the question alongside the image: <Question>

• <Image> Extract documents linked to the question provided in conjunction with the image: <Question>

• <Image> Utilizing the given image, obtain documents that respond to the following question: <Question>

• <Image> Using the given image, access documents that provide insights into the following question: <Question>

• <Image> Obtain documents that correspond to the inquiry alongside the provided image: <Question>

• <Image> With the provided image, gather documents that offer a solution to the question: <Question>

• <Image> Utilizing the given image, obtain documents that respond to the following question: <Question>

\begin{table*}[!hbtp]
    \centering
    \small
    \begin{tabularx}{\textwidth}{XXXXX}
    \toprule
       WIT & IGLUE & KVQA & CC3M & MSMARCO \\
       \midrule
        \includegraphics[valign=c, width=0.19\textwidth]{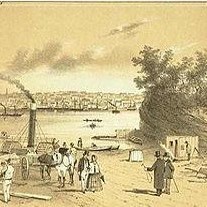}& \includegraphics[valign=c, width=0.19\textwidth]{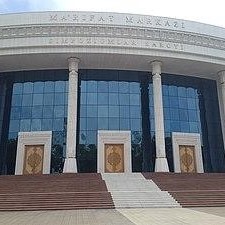} & \includegraphics[valign=c, width=0.19\textwidth]{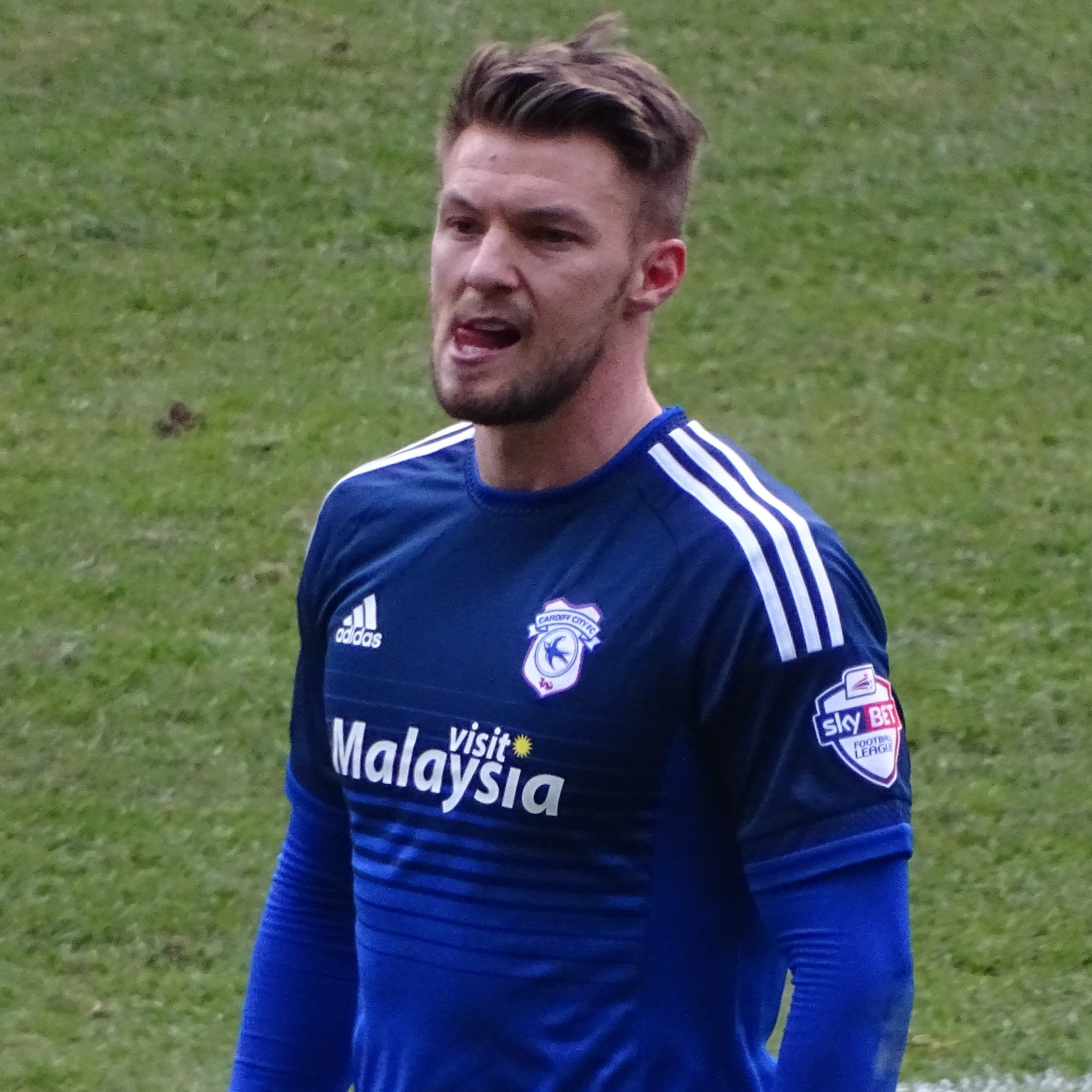} & \includegraphics[valign=c, width=0.19\textwidth]{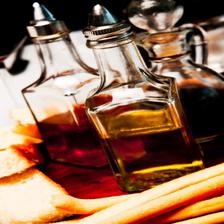} & \includegraphics[valign=c, width=0.19\textwidth]{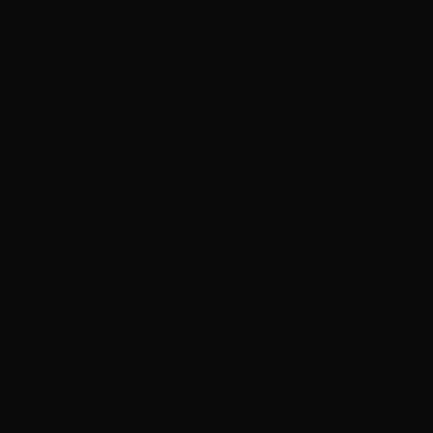} \\
       Describe the image concisely. & Summarize the visual content of the image. & Provide a brief description of the image and the relevant details of the person in the image.& Describe the image concisely. & Retrieve the document that answers this question: how many years did william bradford serve as governor of plymouth colony? \\
       \midrule
        title: PS Herald section title: Formation and operation of the North Shore Steam Company ... &  title: National Library of Uzbekistan hierarchical section title: National Library of Uzbekistan caption ...  & This is an image of Pilkington playing for Cardiff City in 2016. Anthony Pilkington date of birth is ... & olive oil is a healthy ingredient used liberally. & William Bradford (c.1590 - 1657) was an English Separatist leader in Leiden, ... \\
       \midrule
        LLaVA & OVEN & E-VQA & Infoseek & OKVQA \\
        \midrule
       \includegraphics[valign=c, width=0.19\textwidth]{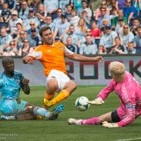}& \includegraphics[valign=c, width=0.19\textwidth]{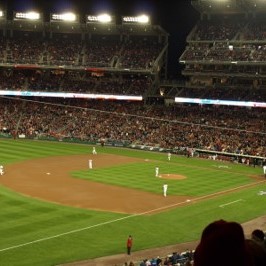} & \includegraphics[valign=c, width=0.19\textwidth]{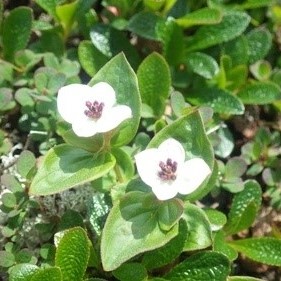} & \includegraphics[valign=c, width=0.19\textwidth]{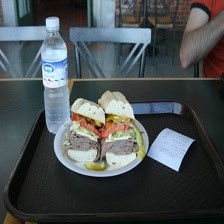} & \includegraphics[valign=c, width=0.19\textwidth]{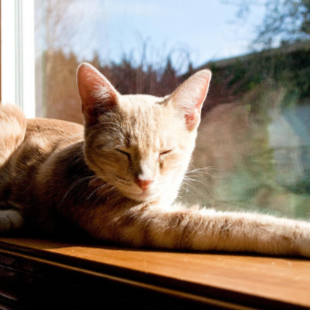} \\
         Provide a brief description of the image along with the following question: what unique situation is occurring in this soccer match? &  Using the provided image, obtain documents that address the subsequent question: what is this park called? & Obtain documents that correspond to
the inquiry alongside the provided image: how big can this plant become? & With the provided image, gather documents that offer a solution to the question: What is the country of origin of this food? & Using the provided image, obtain documents that address the subsequent question: How many teeth does this animal use to have?
 \\
         \midrule 
        In this soccer match, a unique situation is occurring where three men are playing against each other, each wearing a different colored uniform. & Nationals Park is a baseball stadium along the Anacostia River in the Navy Yard neighborhood... & Dwarf cornel is a rhizomatous herbaceous perennial growing to 20cm (8 inches) tall...  & title: Submarine sandwich content: Submarine sandwich A submarine sandwich, also known as a sub...&  Most cats have 26 deciduous teeth and 30 permanent teeth.\\
       \bottomrule
       \end{tabularx}
    \caption{Demonstration of the retrieval tasks for each dataset. We show the image (first row) query, the text query (second row), and the retrieved ground truth document (third row) for each dataset. Since some retrieved documents are too long, we only show part of the document and use ... to stand for continuing documents.}
    \label{tab:dataset_demo}
\end{table*}

\section{Implementation Details}
\subsection{Breakdown of Data Used in Training}
\label{sec:preflmr:appendix:data_proportion}

In Stage 3 Full-scale Fine-tuning, the different sub-tasks in the M2KR dataset are downsampled or duplicated to balance the dataset proportions during training.
The detailed breakdown of the data used in different phases is presented in Table~\ref{tab:preflmr:adjust_data_size}.
We observed that without adjusting the data proportions during training, the model's training losses on certain datasets like WIT, Infoseek, and OVEN decrease much faster than on others once all parameters become trainable. This goes against our goal of training a multi-tasking system. Adjusting the data proportions is crucial to ensure a more consistent learning process across different tasks.

\begin{table}[!ht]
\small
\begin{tabular}{@{}llll@{}}
\toprule
         & Stage 1 & Stage 2 & Stage 3               \\ \midrule
WIT      & 2.8M    & -       & 140K                  \\
IGLUE    & -       & -       & -                     \\
KVQA     & 65K     & -       & 6.5K                  \\
CC3M     & 595K    & -       & 29.8K                 \\
MSMARCO  & 400K    & -       & 40K                   \\
OVEN     & 339K    & -       & 33.9K                 \\
LLaVA    & 351K    & -       & 35.1K                 \\
OKVQA   & 9K      & -       & 90K (repeat 10 times) \\
Infoseek & 100K    & -       & 50K                   \\
E-VQA     & 167K    & 167K    & 167K                  \\ \bottomrule
\end{tabular}
\caption{The dataset sizes are adjusted in Stage 3 in practice.}
\label{tab:preflmr:adjust_data_size}
\end{table}

\subsection{Detailed Hyperparameters}
\label{sec:preflmr:appendix:hyperparameters}

We use the Adam optimizer~\cite{KingmaB14adam} with a fixed learning rate of $10^{-4}$ for the mapping structure and $10^{-5}$ for the rest parameters in all experiments in all training stages.
4 Nvidia A100 GPUs were used with data parallel in all experiments.

Stage 0: Training was run up to 300k steps. The batch size is 8 and the gradient accumulation step is 8. The number of negative examples is 1. The validation ran per 10k steps. The checkpoint was taken at the best Recall@50 on the original MSMARCO validation set, following~\citet{khattab-zaharia-2020-colbert}.
The total training time is approximately 1.5 days per model.

Stage 1: Training was run up to 220k. The batch size is 8 and the gradient accumulation step is 8. The number of negative examples is 4. The validation interval is 10k steps.
The checkpoint was taken at the best Recall@10 on the validation set of WIT in M2KR.
The total training time is approximately 5 days per model.

Stage 2: The intermediate pre-training was run for 12k steps for all experiments. The batch size is 8 and the gradient accumulation step is 8. The number of negative examples is 4.
The total training time is approximately 2 days per model.

Stage 3: Training was run for 50k for all experiments. Training was early-stopped if the performance on WIT or E-VQA decreases for 3 consecutive validation runs. 
Validation was run per 10k steps.
The batch size is 8 and the gradient accumulation step is 8. The number of negative examples is 4.
The total training time is approximately 2 days per model.

Single-task Downstream Fine-tuning: The batch size is 8 and the gradient accumulation step is 8. The number of negative examples is 4. 
For reference, in our experiments, the downstream fine-tuning took 20k, 5k, 1k, 15k, 2.5k steps to achieve the best performance for WIT, OVEN, Infoseek, E-VQA, OKVQA respectively (for the ViT-G + Base-v2 PreFLMR).
The total training time is approximately 1 days per model per task.

VQA Fine-tuning: We used BLIP2-T5XL as the answer generator as in RA-VQAv2~\cite{lin2023finegrained}. The retriever was frozen during training and inference. The batch size is 1 and the gradient accumulation step is 16. 
For each question in a training batch, top-5 relevant documents were pre-extracted using the retriever, and 3 out of 5 were randomly selected.
These 3 documents were concatenated to the question and sent to the answer generator for one forward pass individually.
This setting is to enable training with top-5 documents given limited GPU memory.
The total training time is approximately 2 days per model.

Every model reported in this paper was reproduced once to make sure the training is reproducible.
The best result is reported since the model with the best result will be released to the community.
There is not much difference in the two runs. The absolute difference is less than 0.2 Recall score in most datasets (except that PreFLMR\_ViT-B\_Base-v2 has a -0.4 difference on Infoseek).

\subsection{Large-v1 Training}
\label{sec:preflmr:appendix:large_v1_training}
In our experiments, we found that training Large-v1 during Stage 2/3 was not steady. First, the loss decreased faster than in other systems, like Base-v1, even though Large-v1 had worse system performance. This happened because Large-v1's bigger model capacity made it more prone to overfitting.

Next, the loss suddenly shot up, causing the training to collapse, despite using the same data and strategy as Base-v1. We tried different hyperparameters, like lowering the learning rate to $1e-6$, $3e-6$, but the model still collapsed.

Finally, when we used LoRA~\cite{hu2022lora} with Large-v1 during training, it helped stabilize the process. The LoRA hyperparameters used were: $r=16$, $\alpha=32$, and a dropout rate of 0.05.

\subsection{Model Design in Detail}
\label{sec:preflmr:appendix:model_design}
Similar to FLMR, PreFLMR consists of three components: a vision model $\mathcal{F}_V$, a mapping structure $\mathcal{F}_M$, and a language model $\mathcal{F}_L$.

\paragraph{Feature Extraction.}
The textual query $q$ consists of an instruction and (optionally) a question (e.g., "Utilize the given image to procure documents addressing the following query: [Question]"). We use a language model with hidden size $d_L$ to obtain embeddings for all $N_q$ tokens which are concatenated into matrix $\textbf{Q}_q$:

\begin{equation}
    \begin{aligned}
        \textbf{Q}_q=\mathcal{F}_L(q) \in \mathcal{R}^{N_{q} \times d_L}
    \end{aligned}
    \label{eq:preflmr:text_query_feature}
\end{equation}

Like FLMR, a vision model $\mathcal{F}_V$ encodes the input image $I$,  extracting the [CLS] token embeddings from the last layer. PreFLMR additionally uses the patch embeddings from the penultimate layer of ViT for more complete representation.


\begin{equation}
    \begin{aligned}
        \textbf{Q}_{I, [CLS]}=\mathcal{F}_V(I) \in \mathcal{R}^{1 \times d_V}
    \end{aligned}
    \label{eq:preflmr:cls_feature}
\end{equation}


\begin{equation}
    \begin{aligned}
        \textbf{Q}_{I,PATCH}=\mathcal{F}_{V,-2}(I) \in \mathcal{R}^{N_V \times d_V}
    \end{aligned}
    \label{eq:preflmr:patch_feature}
\end{equation}

The mapping structure $\mathcal{F}_M$ comprises two components: a 2-layer MLP $\mathcal{F}_{M}^{MLP}$ and a Transformer block $\mathcal{F}_{M}^{TR}$.

Following the FLMR model, a 2-layer Multi-Layer Perceptron (MLP) $\mathcal{F}_{M}^{MLP}$ is utilized to convert the initial token embeddings into visual token embeddings with a length of $N_{vt}$ and a hidden size $d_h$:\footnote{Transformation sequence: $\mathcal{R}^{d_V} \rightarrow \mathcal{R}^{N_{vt}d_h/2} \rightarrow \mathcal{R}^{N_{vt}d_h}$, subsequently reshaped into $\mathcal{R}^{N_{vt} \times d_h}$.}

\begin{equation}
\begin{aligned}
\mathbf{Q}_{I}^{MLP} = \mathcal{F}_{M}^{MLP}(\textbf{Q}_{I, [CLS]}) \in \mathcal{R}^{N_{vt} \times d_h}
\end{aligned}
\label{eq:preflmr:mlp_head}
\end{equation}

Moreover, an additional Transformer module $\mathcal{F}_{M}^{TR}$ is introduced to manage all patch embeddings. 
It is a stack of $N_{TR}$ transformer layers with a hidden size $d_L$, followed by a simple MLP layer at the end.
This module leverages cross-attention with the text query $\mathbf{Q}_q$, enabling query-aware image feature mapping.

\begin{equation}
\begin{aligned}
\mathbf{Q}_{I}^{TR} = \mathcal{F}_{M}^{TR}(\mathcal{F}_v(\textbf{Q}_{I,PATCH}), \mathbf{Q}_q) \in \mathcal{R}^{N_{V} \times d_h}
\end{aligned}
\label{eq:preflmr:trans_head}
\end{equation}

Here, $\mathcal{F}v$ represents a 1-layer MLP that adapts the dimension from $d_V$ to $d_L$, which is subsequently transformed to $d_h$ by the linear MLP layer of $\mathcal{F}_{M}^{TR}$. The resultant features from these processes are concatenated to formulate the query embeddings:

\begin{equation}
\begin{aligned}
\mathbf{Q} = \left [\mathbf{Q}_q | \mathbf{Q}_I^{MLP} | \mathbf{Q}_I^{TR} \right ] \in \mathcal{R}^{(N_{vt} + N_{V} + N_q) \times d_h}
\end{aligned}
\label{eq:preflmr:query_emb}
\end{equation}

Furthermore, the document representations in the knowledge base are denoted by $\mathbf{D}$, derived from the document content $d$ with length $l_D$:
\begin{equation}
\begin{aligned}
    \mathbf{D} = \mathcal{F}_l(\mathcal{F}_L(d)) \in \mathcal{R}^{l_D\times d_h},
\end{aligned}
\label{eq:preflmr:item}
\end{equation}

where $\mathcal{F}_l$ signifies a straightforward MLP layer tasked with mapping $d_L$ to $d_h$, thereby aligning the dimensionality with the query embeddings.

\textbf{Multi-Modal Late Interaction.}
The relevance score between a question-image pair $\bar{\mathbf{q}}=(q, I)$ and a document $d$ is calculated using a late-interaction paradigm:

\begin{equation}
r(\bar{\mathbf{q}}, d) = {r}((q, I), d) = \sum_{i=1}^{l_Q} \max_{j=1}^{l_D} \mathbf{Q}_{i} \mathbf{D}_{j}^\top
\label{eqn:flmr:late_interaction}
\end{equation}

where $l_Q=N_{vt} + N_{V} + N_q$.
For each token in the query, the system aggregates the maximum relevance score across all tokens in the document.

\textbf{Training and Inference.}
For model training, documents $d^*$ corresponding to a query $q$ are considered gold (positive) samples. We incorporate random negative sampling from the corpus.\footnote{In multi-dataset scenarios, negative samples are selected from the same corpus as $d^*$.} Additionally, we adopt in-batch negative sampling as suggested by~\citet{karpukhin-etal-2020-dense}, treating all non-corresponding documents in a batch as negatives for $q$, denoted as $\mathcal N(q)$. The model is trained using a contrastive loss across the dataset $\mathcal{D}$:

\begin{equation}
   \mathcal{L} = - \sum_{(q, d^*)\in \mathcal{D}} \log{\frac{\exp{\left(r(q,d^*)\right)}}{\exp{\left(r(q, d^*)\right)}+\hspace*{-2ex}\displaystyle\sum_{z\in \mathcal{N}(q)}\hspace*{-1ex}\exp{\left(r(q,z)\right)}}}
   \label{eq:flmr:dprloss}
\end{equation}

Post-training, all documents are indexed through PLAID~\citep{keshav2022plaid} for efficient late-interaction retrieval. For detailed evaluation of retrieval efficiency, we refer readers to \citet{lin2023finegrained}.

\section{Ablation Study on Pre-training Stages}

\begin{table*}[!h]
\resizebox{\textwidth}{!}{%
\begin{tabular}{@{}lccccccccc@{}}
\toprule
Model          & WIT  & IGLUE & KVQA & MM   & OVEN & LLaVA & Infoseek & E-VQA & OKVQA \\ \midrule
PreFLMR\_ViT-B\_Base-v1 & 41.7 & 57.3  & 28.6 & 79.5 & 46.3 & 67.2  & 48.8     & 67.9  & 66.1  \\
~~\textit{w/o Stage 0}    & 25.5 & 28.8  & 21.0 & 56.5 & 33.9 & 55.0  & 42.5     & 51.8  & 64.5  \\
~~\textit{w/o Stage 1}    & 38.2 & 54.9  & 26.6 & 78.0 & 45.5 & 62.8  & 44.6     & 61.9  & 65.5  \\
~~\textit{w/o Stage 2}    & 41.2 & 56.8  & 26.5 & 78.2 & 43.7 & 65.0  & 47.0     & 57.3  & 65.1  \\ \bottomrule
\end{tabular}%
}
\caption{Retrieval performance when disabling pre-training stages. Removal of any stage deteriorated the performance.}
\label{tab:preflmr:ablate_pretraining_stages}
\end{table*}

We present the ablation study for the four pre-training stages in Table~\ref{tab:preflmr:ablate_pretraining_stages}. To ensure consistent comparison, these ablated versions underwent the same number of training steps as PreFLMR\_ViT-B\_Base-v2. The results clearly indicate that the removal of any stage deteriorates performance. Specifically, disabling Stage 0 (i.e. using untrained text encoder) leads to the most significant performance decline because the text encoder is not pre-trained on late-interaction, resulting in a diminished ability to capture fine-grained relevance within the same computational budget. Note that removing Stage 0 leads to collapsed performance on Stage 1, where the text encoder is frozen.
Furthermore, removing Stage 2 notably affects the performance on E-VQA more than on other KB-VQA datasets, highlighting the challenge posed by E-VQA and the necessity of intermediate pre-training.

\section{V-Entropy-based Analysis of Intermediate Pre-training}
\label{sec:preflmr:appendix:v_information}



V-Entropy~\cite{VInformation}, $H_\mathcal{V}(Y|X)$, is the minimal Negative Log-Likelihood (NLL) achievable by the probabilistic predictor $f(Y|X)$ under the predicative family $\mathcal{V}$. A predicative family can be viewed as the set of reachable models under a certain model architecture and training budgets.

We define Mutual Information $I_{\mathcal{V}[N_f]}(D_1 \rightarrow D_2)$ between datasets $D_1$ and $D_2$ in Eq.\ref{eq:dataset_MI-def}.
We define $H_{\mathcal{V}[N_f]}(D_2)$ as the minimal achieved NLL loss on the validation set of dataset $D_2$ after $N_f$ training steps on $D_2$.
$\mathcal{V}[N_f, D_1, N_t]$ denotes the set of reachable models after $N_f$ fine-tuning steps on $D_2$ starting from a checkpoint that has been trained on dataset $D_1$ for $N_t$ steps. 
This is V-Entropy with additional predictive family specification. 

\begin{equation}
\begin{aligned}
    I_{\mathcal{V}[N_{f}]} (D_1 \rightarrow D_2) &= H_{\mathcal{V}[N_{f}]}(D_2) \\
    &\quad - H_{\mathcal{V}{[N_{f}, D_{1}, N_{t}]}}(D_2)
\end{aligned}
\label{eq:dataset_MI-def}
\end{equation}

Intuitively, $D_1$ has high mutual information with $D_2$ if models initialized from $D_1$ checkpoints attain much lower NLL loss compared to models initialized without training on $D_1$. $N_f$ and $N_t$ set the computation constraints for training on $D_2$ and $D_1$, respectively. 
In our experiment, $\mathcal{V}$ is the PreFLMR architecture, $D_1$ is the E-VQA dataset and $D_2$ is the training set of M2KR. 
$N_f$ corresponds to $N_{inter}$ in Sec.~\ref{sec:Analysis of Intermediate Pre-training}, which is the intermediate training steps on the E-VQA dataset.
In the analysis, we set $N_f$ to 5,000 and sweep $N_t$ from 0 to 25,000 in intervals of 5,000.  

We refer readers to \citet{VInformation} for detailed properties of V-Entropy and emphasize that $I_{\mathcal{V}[N_f]}(D_1 \rightarrow D_2)$ is an empirical value we define to estimate mutual information between datasets. It is different from the V-Information defined in \citet{VInformation} which estimates the mutual information between model input and output.

\section{Qualitative Analysis for OKVQA and E-VQA}
\label{sec:preflmr:comparing_okvqa_evqa}
In this section, we compare examples from the OKVQA and E-VQA datasets to highlight their differences. To avoid cherry-picking, we use examples from its official website\footnote{\href{https://okvqa.allenai.org/}{https://okvqa.allenai.org/}} for OKVQA. Similarly, we use the examples included in the paper for E-VQA. Table~\ref{tab:demo_okvqaevqa} presents three examples from each dataset. 

The OKVQA examples typically require common sense knowledge, like `people attend church on Sundays' or `firetrucks use fire hydrants.' State-of-the-art Large Language Models (LLMs) often have this common sense knowledge inherently built-in, making additional knowledge retrieval less impactful for OKVQA tasks.


In contrast, E-VQA examples demand more specialized, expert-level knowledge, necessitating an effective knowledge retrieval system. For instance, correctly answering a question about 'Acacia paradoxa' requires first retrieving the relevant document providing specific information about this plant species. Enhancing the knowledge retrieval system to source accurate documents is crucial for improving performance on the E-VQA dataset.

\begin{table*}[!htp]
    \centering
    \small
    \begin{tabularx}{\textwidth}{XXX}
    \toprule
    \multicolumn{3}{c}{OKVQA}\\
    \midrule
       \includegraphics[valign=c, width=0.31\textwidth]{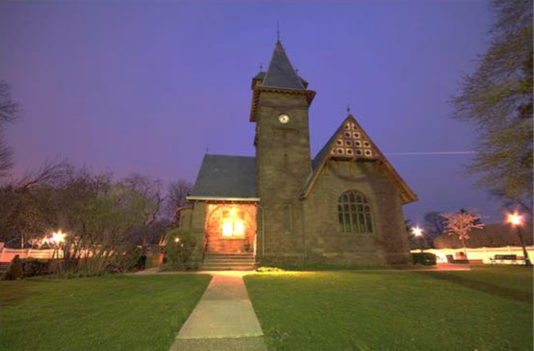} & \includegraphics[valign=c, width=0.31\textwidth]{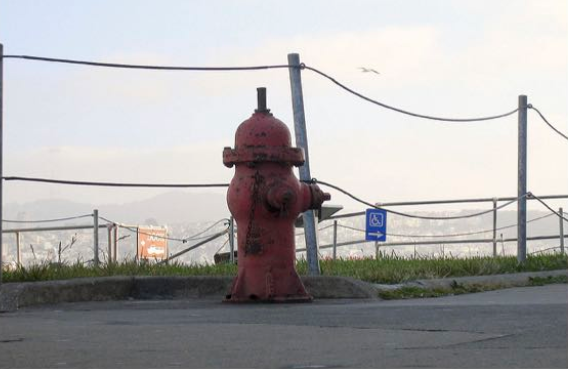} & \includegraphics[valign=c, width=0.31\textwidth]{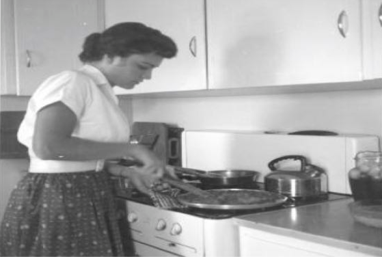}  \\
        Q: What days might I most commonly go to this building? &Q: What sort of vehicle uses this item? & Q: Is this photo from the 50's or the 90's?\\
        A: Sunday &A: firetruck & A: 50's \\
        \midrule 
        \multicolumn{3}{c}{E-VQA}\\
        \midrule
        \includegraphics[valign=c, width=0.31\textwidth]{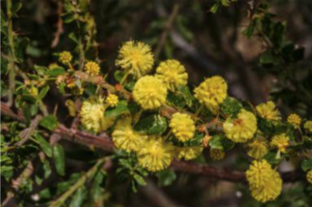} & \includegraphics[valign=c, width=0.31\textwidth]{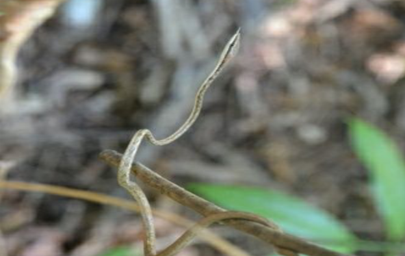} & \includegraphics[valign=c, width=0.31\textwidth]{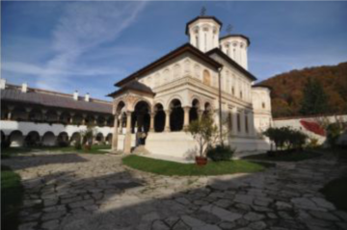}   \\
        Q: How many feet tall does this tree grow to? &Q: How many eggs does this reptile typically lay? & Q: Who founded this monastery? \\
        A: 7 to 13 &A: 3-6 & A: Prince Constantin Brâncoveanu \\
    \bottomrule
    \end{tabularx}
    
    \caption{Demonstrative examples from OKVQA and E-VQA. Questions in E-VQA require more domain knowledge to answer generally.}
    \label{tab:demo_okvqaevqa}
\end{table*}

\section{Artifacts and License}
We list the resources used and their License below:

(1) huggingface-transformers (Apache License 2.0) provides pre-trained model checkpoints for BLIP 2, DPR, and their tokenizers: \hyperlink{https://github.com/huggingface/transformers}{https://github.com/huggingface/transformers}

(2) FAISS~\citep{johnson2019billion} (MIT License) is used to index document embeddings for fast retrieval with DPR: \hyperlink{https://github.com/facebookresearch/faiss}{https://github.com/facebookresearch/faiss}

(3) huggingface-PEFT (Apache License 2.0) for parameter-efficient LoRA fine-tuning: \hyperlink{https://github.com/huggingface/peft}{https://github.com/huggingface/peft}

(4) PLAID and ColBERTv2 (MIT License): \hyperlink{https://github.com/stanford-futuredata/ColBERT}{https://github.com/stanford-futuredata/ColBERT}

(5) RA-VQA-v2 official repository with training and testing codes (GNU General Public License v3.0): \hyperlink{https://github.com/LinWeizheDragon/Retrieval-Augmented-Visual-Question-Answering}{https://github.com/LinWeizheDragon/Retrieval-Augmented-Visual-Question-Answering}. 

(6) Datasets used in building the M2KR benchmark: 
\begin{itemize}[leftmargin=5.5mm]
\setlength\itemsep{-0.3em}
    \item WIT (Creative Commons Attribution-ShareAlike 3.0 Unported \href{https://github.com/google-research-datasets/wit/blob/main/LICENSE}{https://github.com/google-research-datasets/wit/blob/main/LICENSE});
    \item MSMARCO (non-commercial research purposes only~\href{https://microsoft.github.io/msmarco/}{https://microsoft.github.io/msmarco/});
    \item CC3M (Free for any purposes~\href{https://github.com/google-research-datasets/conceptual-captions}{https://github.com/google-research-datasets/conceptual-captions});
    \item LLaVA, the image of LLaVA is a subset of CC3M. It should inherit the license of CC3M. The conversation data follows policy of OpenAI: \href{https://openai.com/policies/terms-of-use}{https://openai.com/policies/terms-of-use}.
    \item IGLUE (MIT license \href{https://github.com/e-bug/iglue/blob/main/LICENSE}{https://github.com/e-bug/iglue/blob/main/LICENSE});
    \item KVQA (No specific license is mentioned~\href{https://malllabiisc.github.io/resources/kvqa/}{https://malllabiisc.github.io/resources/kvqa/});
    \item OVEN (Apache-2.0 license \href{https://github.com/open-vision-language/oven/blob/main/LICENSE}{https://github.com/open-vision-language/oven/blob/main/LICENSE});
    \item E-VQA (no specific license mentioned \href{https://github.com/google-research/google-research/tree/master/encyclopedic_vqa}{https://github.com/google-research/google-research/tree/master/encyclopedic\_vqa});
    \item Infoseek (Apache License 2.0 \href{https://github.com/open-vision-language/infoseek/blob/main/LICENSE}{https://github.com/open-vision-language/infoseek/blob/main/LICENSE})
    \item OKVQA (Copyright (c) 2021, Chen Qu and Center for Intelligent Information Retrieval, University of Massachusetts, Amherst. \href{https://github.com/prdwb/okvqa-release/blob/main/LICENSE}{https://github.com/prdwb/okvqa-release/blob/main/LICENSE})
\end{itemize}

In particular, we emphasize that no changes are made to the original data of all the datasets used in our work. Our released models and artifacts should only be used for non-commercial purposes. By using the pre-trained models, users agree to respect the terms and conditions of the datasets used in pre-training.

\section{PreFLMR model performance radar chart on M2KR tasks}

\begin{figure*}[!h]
    \centering
    \includegraphics[width=1\linewidth]{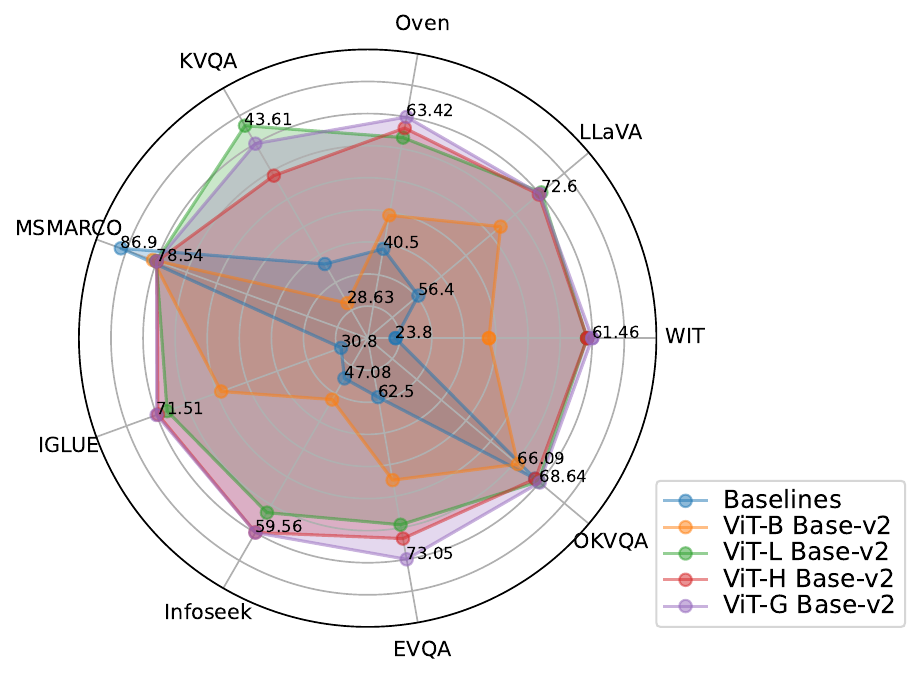}
    \caption{PreFLMR achieves strong performance on the M2KR benchmark. The scale of the plot is adjusted for better visualization. The best and worst numbers of each task are annotated.
    }
    \label{fig:preflmr:radar_plot}
\end{figure*}

Fig.~\ref{fig:preflmr:radar_plot} demonstrates the performance of PreFLMR with a radar plot. The best and worst numbers of each task are annotated.

\section{AI Assistance}

Our coding work was assisted by Github Copilot.\footnote{\href{https://github.com/features/copilot}{https://github.com/features/copilot}}
OpenAI ChatGPT\footnote{\href{https://chat.openai.com/}{https://chat.openai.com/}} was only used in proofreading and spell-checking. We claim that the content presented in this paper was fully original.

\end{document}